\documentclass[10pt,twocolumn,letterpaper]{article}
\usepackage{cvpr}
\usepackage{times}
\usepackage{epsfig}
\usepackage{graphicx}
\usepackage{amsmath}
\usepackage{amssymb}
\usepackage{booktabs}
\usepackage{multirow}
\usepackage{caption}

\usepackage[ruled,vlined,linesnumbered]{algorithm2e}
\DontPrintSemicolon
\SetKwInput{KwIn}{Input}
\SetKwInput{KwOut}{Output}
\SetKw{KwInit}{Init}

\usepackage{xcolor}
\definecolor{cvprblue}{rgb}{0.21,0.49,0.74}

\usepackage[pagebackref,breaklinks,colorlinks,allcolors=cvprblue]{hyperref}

\def\confName{CVPR}
\def\confYear{2026}

\begin{document}

\title{\textsc{SHands}: A Multi-View Dataset and Benchmark for Surgical Hand-Gesture and Error Recognition Toward Medical Training}

\author{
Le~Ma$^{1}$ \quad
Thiago~Freitas~dos~Santos$^{1}$ \quad
Nadia~Magnenat-Thalmann$^{1}$ \quad
Katarzyna~Wac$^{2}$\\[0.5em]
$^{1}$MIRALab, $^{2}$Quality of Life Technologies Lab (QoL Lab), University of Geneva\\
{\tt\small le.ma@unige.ch, thiago.freitas@miralab.ch, thalmann@miralab.ch, katarzyna.wac@unige.ch}
}

\maketitle
\thispagestyle{empty}

\begin{figure*}[t]
  \centering
  \includegraphics[width=0.86\linewidth]{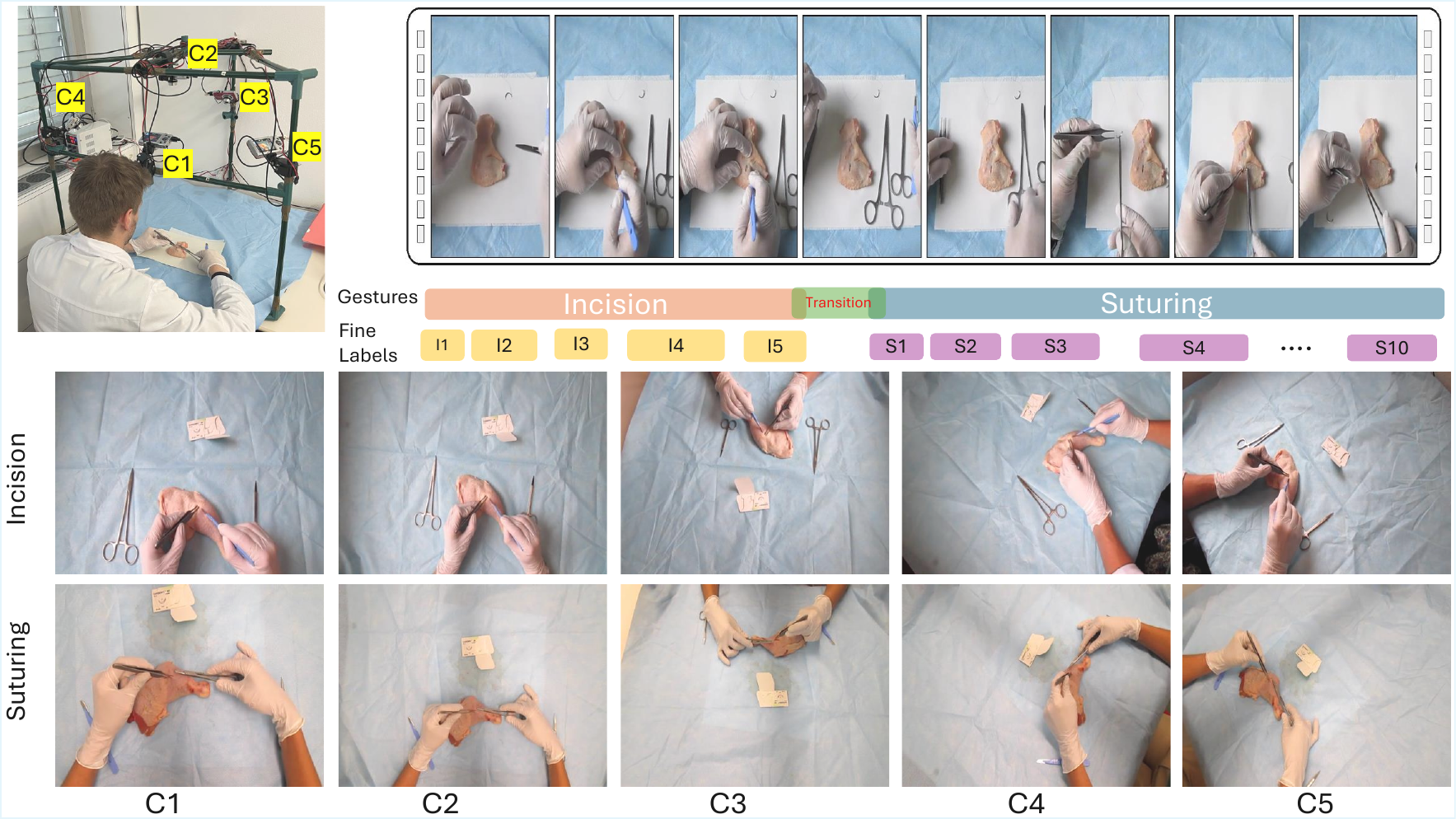}
  \caption{The \textsc{SHands} multi-view dataset. A five-camera RGB setup (C1–C5) records synchronized multi-view videos of incision and suturing tasks on ex vivo tissue. The top row shows a gesture-annotated timeline with fine-grained labels (I1–I5 for incision, S1–S10 for suturing) and transition boundaries. The bottom rows show aligned frames from all views for both incision and suturing, highlighting the complementary spatial information captured across cameras.}
  \label{fig:overview}
\end{figure*}

\begin{abstract}
In surgical training for medical students, proficiency development relies on expert-led skill assessment, which is costly, time-limited, difficult to scale, and its expertise remains confined to institutions with available specialists. Automated AI-based assessment offers a viable alternative, but progress is constrained by the lack of datasets containing realistic trainee errors and the multi-view variability needed to train robust computer vision approaches. To address this gap, we present Surgical-Hands (\textsc{SHands}), a large-scale multi-view video dataset for surgical hand-gesture and error recognition for medical training. \textsc{SHands} captures linear incision and suturing using five RGB cameras from complementary viewpoints, performed by 52 participants (20 experts and 32 trainees) each completing three standardized trials per procedure. The videos are annotated at the frame level with 15 gesture primitives and include a validated taxonomy of 8 trainee error types, enabling both gesture recognition and error detection. We further define standardized evaluation protocols for single-view, multi-view, and cross-view generalization, and benchmark state-of-the-art deep learning models on the dataset. \textsc{SHands} is \href{https://LM-collab.github.io/Shands-project-page/}{publicly released} to support the development of robust and scalable AI systems for surgical training grounded in clinically curated domain knowledge. 
\end{abstract}

\section{Introduction}
\label{sec:introduction}
Surgical proficiency depends on the ability to execute precise hand movements and handle instruments correctly to prevent technical errors~\cite{martin1997objective,shayan2023measuring}. Traditionally, such proficiency is assessed through expert observation~\cite{olsen2022crowdsourced,olsen2025untangling}, a process that is expensive, time-intensive, and limited in scalability, especially in surgical education, where hands-on training is essential. Recent advances in computer vision and AI promise scalable and objective skill evaluation~\cite{pedrett2023technical,ma2024transsg,power2025automated}. However, these methods remain underdeveloped due to a lack of datasets capturing realistic surgical errors and the variability in hand motion across viewing angles. 

Existing surgical video datasets can be categorized into two main groups, each addressing only part of the skill assessment challenge. Robotic-surgery datasets such as JIGSAWS~\cite{gao2014jhu} provide synchronized video and kinematic streams but fail to capture the manual hand–tool coordination central to open surgical training, where dexterity develops without robotic mediation. Endoscopic phase datasets, such as Cholec80~\cite{twinanda2017endonet} and M2CAI16~\cite{jin2018sv}, offer single-view laparoscopic recordings but lack the fine-grained gesture boundaries and clinically validated error labels needed for targeted feedback. Moreover, because multi-view recordings of open procedures are largely absent from these datasets, insights from outside the domain help motivate their value. Multi-view action recognition datasets like NTU RGB+D~\cite{shahroudy2016ntu} and Assembly101~\cite{sener2022assembly101} demonstrate the value of synchronized viewpoints for modeling complex manipulations. Yet, they provide no clinical specificity, expert supervision, or taxonomy of surgical errors.

To address these limitations, we introduce Surgical-Hands (\textsc{SHands}), a large-scale, multi-view RGB video dataset for open surgical hand-gesture and error recognition. As illustrated in Figure~\ref{fig:overview}, the dataset was recorded using five synchronized cameras capturing complementary viewpoints at 25 frames per second with a resolution of 640×480 pixels, substantially reducing occlusion and enabling cross-view inference. The dataset includes recordings of 52 participants (20 expert surgeons and 32 medical trainees), each performing standardized incision and suturing procedures in three independent trials. Every frame is annotated with 15 gesture primitives representing the fundamental components of surgical action, and with a clinically validated taxonomy of eight error types, including improper grip, incorrect trajectory, tissue damage, and insufficient tension. Data were collected using ex vivo chicken tissue to ensure both realistic tool–tissue interactions and standardization across sessions.

We benchmark several state-of-the-art video recognition architectures, including VideoMAE~\cite{tong2022videomae} and TimeSformer~\cite{bertasius2021timesformer}, under standardized protocols for single-view, multi-view, and cross-view generalization. Our experiments demonstrate that integrating multi-view information significantly enhances performance for both gesture recognition and error detection, underscoring the dataset’s value for advancing solutions in AI-driven surgical training.

In summary, our main contributions are as follows:
\begin{itemize}\setlength{\itemsep}{-4.6 pt}
    \item We introduce \textsc{SHands}, to the best of our knowledge, the first multi-view RGB dataset for open surgical hand-gesture and error recognition, captured from five synchronized viewpoints and involving 52 participants across expert and trainee skill levels.
    \item We provide fine-grained, frame-level annotations for 15 surgical gesture primitives and an eight-category error taxonomy, including synchronized label propagation to support single- and multi-view learning. 
    \item We establish comprehensive benchmarks using state-of-the-art video backbones and multi-view fusion methods, providing standardized evaluation protocols for future research on robust, scalable AI-based surgical skill assessment.
\end{itemize}

\begin{table*}[t]
\centering
\small
\setlength{\tabcolsep}{4pt}
\begin{tabular}{l c c c c c c l}
\toprule
\textbf{Dataset} & \textbf{Hours} & \textbf{Participants} & \textbf{Views} & \textbf{Classes} & \textbf{Errors} & \textbf{Avg. V. Length (s)} & \textbf{Domain} \\
\midrule
\multicolumn{8}{l}{\textit{Surgical Datasets}} \\
JIGSAWS\cite{gao2014jhu} & 3.5 & 8 & 1 & 15 & -- & 120 & Robotic bench-top \\
DESK\cite{madapana2019desk} & 5.2 & 11 & 1 & 7 & -- & - & Robotic bench-top \\
Cholec80\cite{twinanda2017endonet} & 50 & 13 & 1 & Phases only & -- & \textbf{2300} & Endoscopic OR \\
M2CAI16\cite{jin2018sv} & 16 & - & 1 & Phases only & -- & - & Endoscopic OR \\
CATARACTS\cite{alhajj2019cataracts} & 9 & 50 & 1 & Steps/Tools & -- & 656 & Microsurgery \\
MVOR \cite{Srivastav2018MVOR} & -- & - & 3 & Pose/Boxes & -- & -- & OR multi-view \\
EgoSurgery\cite{fujii2024egosurgery} & 15 & 8 & 1 & Phases/Tool Boxes & -- & -- & Open Surgery \\
\midrule
\multicolumn{8}{l}{\textit{Multi-View Action Recognition Datasets}} \\
NTU RGB+D 120 \cite{liu2019ntu} & ~200 & 106 & 3 & 120 actions & -- & - & General actions \\
PKU-MMD\cite{liu2017pku}  & 50 & 66 & 3 & 51 actions & -- & 3.8 & General actions \\
Assembly101\cite{sener2022assembly101} & 513 & 53 & \textbf{8} & 101 coarse/1380 & -- & 426 & Assembly tasks \\
\midrule
\multicolumn{8}{l}{\textit{Action Quality Assessment Datasets} player videos} \\
MTL-AQA\cite{Parmar2019MTLAQA} & 1.5 & 1412 & 1 & Dive scores & -- & 4.1 & Sports (diving) \\
FineDiving \cite{Xu2022FineDiving}& 3.5 & 3000  & 1 & Dive scores & -- & 4.2 & Sports (diving) \\
BASKET\cite{Pan2025BASKET} & \textbf{4477} & 32,232  & 1 & 20 skills & -- & 500 & Sports (basketball) \\
\midrule

\textbf{SHands (Ours)} & 10 & 52 & 5 & \textbf{2 tasks, 15+8} & \textbf{8} & 660 & \textbf{Open-Surgery Hands} \\
\bottomrule
\end{tabular}
\caption{\textbf{Comparison with existing surgical and multi-view action datasets.} 
Our dataset is the \textbf{first} to provide synchronized multi-view open hand surgery videos with comprehensive gesture and error annotations. 
}
\label{tab:dataset_comparison}
\end{table*}

\section{Related Work}
This section reviews prior research most relevant to our work. We organize it into three directions: (i) surgical video datasets that have enabled automatic skill assessment and tool tracking; (ii) multi-view action recognition studies demonstrating the importance of viewpoint diversity in understanding fine-grained gestures; and (iii) action recognition methods that form the algorithmic foundation for gesture segmentation and temporal modeling. Together, these lines of research highlight the gap of multi-view, clinically validated datasets for open-surgical gesture and error recognition, which motivates the creation of \textsc{SHands}.

\subsection{Surgical Video Datasets}

Datasets in robotic surgery, such as JIGSAWS~\cite{gao2014jhu}, introduced synchronized kinematic and video data for bench-top tasks, followed by DESK~\cite{madapana2019desk} and ROSMA~\cite{rivas2023surgical}, which broadened the coverage of robotic tasks. However, these datasets capture robot-mediated interactions rather than manual hand–tool coordination. 

Works also center on endoscopic video analysis. Datasets including Cholec80~\cite{twinanda2017endonet}, M2CAI16~\cite{jin2018sv}, CATARACTS~\cite{alhajj2019cataracts}, and AutoLaparo~\cite{wang2022autolaparo} have advanced phase recognition, tool detection, and workflow segmentation. However, their single-camera laparoscopic setting omits direct visualization of the surgeons’ hands and instruments. Broader operating room datasets such as MM-OR~\cite{ozsoy2025_mmor}, CPR-Coach~\cite{wang2024cpr}, and 4D-OR~\cite{ozsoy2022_4dor} model complex team dynamics using multimodal sensors, but their emphasis on holistic scene understanding diverges from our goal of detailed manual technique evaluation.

Further specialized datasets (AVOS~\cite{sharghi2020operatingroom}, C2D2~\cite{malpani2016systemevents}, and Surgical-VQLA~\cite{bai2023surgicalvqla}) investigate tool tracking, crowd-sourced skill annotation, and video question answering. Yet these datasets typically lack frame-level gesture boundaries and systematic error taxonomies, both of which are critical for generating automated feedback in training systems.

Table~\ref{tab:dataset_comparison} provides a consolidated comparison of these datasets alongside multi-view action-recognition and action-quality assessment benchmarks. The overview highlights a consistent pattern: surgical datasets offer clinical realism but remain predominantly single-view or lack systematic error annotation. In contrast, multi-view action datasets provide complementary strengths but no surgical supervision. \textsc{SHands} is the first to unify multi-view capture, fine-grained gestures, and clinically validated error labeling for open manual surgery.

\subsection{Multi-View Action Recognition}
Outside the surgical domain, extensive research has explored how multi-view capture enhances the recognition of complex human activities. Large-scale benchmarks such as NTU RGB+D~\cite{shahroudy2016ntu,liu2019ntu} and PKU-MMD~\cite{liu2017pku} established cross-view protocols using RGB-D sensors to evaluate viewpoint invariance. For fine-grained manipulation, H2O~\cite{kwon2021h2o} and Assembly101~\cite{sener2022assembly101} focus on hand–object interactions across synchronized views, offering valuable insight into tool usage and coordination. Exploring extended procedural contexts: Ego-Exo4D~\cite{grauman2024egoexo4d} pairs egocentric and exocentric recordings to study complementary perspectives, while EgoExoLearn~\cite{huang2024egoexolearn} addresses weakly supervised alignment between them. Moreover, instructional video corpora such as ATA~\cite{ghoddoosian2023ata}, HoloAssist~\cite{wang2023holoassist}, LOGO~\cite{zhang2023logo}, and VidChapters-7M~\cite{yang2023vidchapters7m} facilitate segmentation and procedural reasoning in non-clinical settings.

These studies collectively demonstrate that viewpoint diversity enhances robustness and contextual understanding in action recognition. Motivated by this, we extend the multi-view paradigm to surgical gesture and error recognition, integrating diverse camera capture with clinically validated annotations. The first, to the best of our knowledge, to apply this principle in surgical skill assessment.

\subsection{Action Recognition Methods}
Increasingly powerful architectures drive progress in video understanding. Transformer-based and hierarchical backbones—SlowFast~\cite{feichtenhofer2019slowfast}, MViT~\cite{fan2021mvit}, TimeSformer~\cite{bertasius2021timesformer}, and ViViT~\cite{arnab2021vivit}—have achieved strong performance on generic datasets. Self-supervised and masked pretraining frameworks such as VideoMAE~\cite{tong2022videomae,wang2023videomaev2,hu2024multi} further improve representation quality with reduced supervision. Specifically in the surgical domain, specialized pretraining approaches such as Endo-FM~\cite{wang2023endofm}, EndoViT~\cite{batic2024endovit}, and SurgMAE~\cite{jamal2023surgmae} tailor these architectures for endoscopic imagery but do not address multi-camera open-surgery cases.

Multi-view methods extend these architectures to learn view-invariant or view-fused representations. For instance, DVANet~\cite{siddiqui2024dvanet} disentangles view and action features through contrastive learning, and M$^3$Net~\cite{tang2023m3net} fuses embeddings across views for few-shot recognition. Self-supervised variants such as CVRL~\cite{qian2021cvrl} and MV2-MAE~\cite{wang2024mv2mae} leverage cross-view reconstruction, while HCTransformer~\cite{lin2024human} and graph-based reasoning models~\cite{ho2025dejavid,lin2025mv} incorporate human priors to enhance viewpoint robustness.

For temporal gesture segmentation, architectures like MS-TCN~\cite{farha2019ms}, MS-TCN++~\cite{li2020mstcnpp}, and ASFormer~\cite{yi2021asformer} refine predictions through multi-stage temporal modeling. Within surgical applications, transformer-based gesture recognition and phase understanding models, e.g., SKiT~\cite{liu2023skit}, LoViT~\cite{liu2025lovit}, multimodal fusion approaches~\cite{weerasinghe2024multimodal}, achieve state-of-the-art results on JIGSAWS and Cholec80, yet remain limited to single-view laparoscopic footage.

\section{The Surgical Hand-Gesture (\textsc{SHands}) Dataset}

Despite these advances, current models remain constrained by the limitations of existing datasets. Specifically, their reliance on single-view video and the absence of error labels or open-surgery hand–tool interactions. To address these gaps, we introduce \textsc{SHands}, a multi-view video dataset designed for fine-grained surgical hand-gesture recognition and clinically validated error detection in open-surgery training. It captures two foundational manual skills (incision and suturing) performed on ex vivo chicken tissue, providing realistic tool–tissue interactions under standardized and reproducible conditions.

\subsection{Multi-View Capture System}
As depicted in Figure~\ref{fig:overview}, the capture rig comprises five static RGB cameras (Canon PowerShot A2500; 25 fps, $640{\times}480$ px) mounted on a rigid frame surrounding the surgical workspace. These cameras are positioned to provide complementary viewpoints, combining top-down and oblique angles to minimize occlusions and facilitate cross-view inference. Frame-accurate temporal alignment is guaranteed through \emph{hardware-level synchronization} using the Canon Hack Development Kit (CHDK). Supported by a measurable verification procedure, this synchronization mechanism eliminates the need for post-hoc registration and enables direct pixel-level correspondence across all views. Ultimately, this setup produces a rich and multi-perspective representation of the procedure, effectively capturing the fine nuances of hand–tool–tissue interactions.

\subsection{Participants and Experiment Protocol}
The dataset comprises 52 participants: 20 certified surgeons with clinical experience and 32 medical trainees at different training stages. This composition provides natural variability in performance and technique, which is essential for developing models that generalize across skill levels. Before data collection, participants provided written informed consent under an ethical protocol approved by the University of Geneva's Ethical Committee. To ensure anonymity, only participants' hands and forearms were recorded.

For the medical students, each session followed a standardized training and evaluation protocol. Before the recording began, students first watched a reference video showing an expert performing the target procedures: linear incision and suturing. A trained member of the research team then provided verbal guidance, following the instructions of a medical professional, explaining the correct sequence of gestures, proper instrument handling, and the expected surgical technique.

After this brief instructional phase, students independently performed both procedures, repeating each three times to capture performance variability and enable consistent evaluation. The expert surgeons followed the same repetition protocol—three trials per procedure—without the pre-training step. All trials were carried out at each participant’s self-determined pace, without external intervention, to preserve natural execution variability rather than constrained or scripted motions.

This experiment protocol aims to ensure that \textsc{SHands} reflects the realistic range of surgical performance, from trainee learning behaviors to expert fluency, under ethically compliant and reproducible conditions.

\subsection{Gesture and Error Annotation}



\textsc{SHands} introduces a two-level annotation taxonomy, developed in collaboration with surgical educators, to support gesture recognition and error detection. This hierarchical scheme comprises: (1) 15 Gesture Primitives(Table~\ref{tab:correct_gestures}), which decompose incision (I1–I5) and suturing (S1–S10) into clinically meaningful units with frame-level temporal boundaries; and (2) 8 Error Categories(Table~\ref{tab:incorrect_gestures}), reflecting expert-validated trainee mistake criteria. For suturing, trials are standardized to require the completion of three knots. To ensure temporal completeness, a Background/Idle class is defined for segments containing no meaningful gestures or unlabeled motions. Annotations were performed by a single annotator (guided by an educator's definitions) and verified by an expert surgeon to ensure label consistency. This dual-layered approach facilitates research in multi-view fusion, temporal segmentation, and expert-interpretable error recognition.

\begin{table}[t]
\centering
\small
\begin{tabular}{lll}
\toprule
\textbf{Task} & \textbf{ID} & \textbf{Description} \\
\midrule
\multirow{5}{*}{Incision}
 & I1  & Grasping scalpel and forceps \\
 & I2  & Positioning toward tissue site \\
 & I3  & Stabilizing surrounding tissue \\
 & I4  & Executing incision with blade \\
 & I5  & Retracting incised tissue \\
\midrule
\multirow{10}{*}{Suturing}
 & S1  & Grasping forceps and needle holder \\
 & S2  & Securing needle in holder \\
 & S3  & Positioning toward tissue site \\
 & S4  & Elevating wound edge \\
 & S5  & Passing needle through tissue \\
 & S6  & Pulling suture with needle holder \\
 & S7  & Making a knot \\
 & S8  & Drop the needle holder, grasp a scissors \\
 & S9  & Cutting excess suture \\
 & S10 & Re-grasping needle holder \\
\bottomrule
\end{tabular}
\caption{Surgical gesture taxonomy for incision and suturing. Each gesture represents a clinically meaningful primitive motion used in open-surgery training.}
\label{tab:correct_gestures}
\end{table}

\begin{table}[t]
\centering

\small
\begin{tabular}{llp{5cm}}
\toprule
\textbf{Task} & \textbf{ID} & \textbf{Error Description} \\
\midrule
\multirow{3}{*}{Incision}
 & II1 & Improper instrument grip \\
 & II2 & Incorrect angle \\
 & II3 & Repetitive cutting \\
\midrule
\multirow{5}{*}{Suturing}
 & IS1 & Incorrect needle handling \\
 & IS2 & Excessive force application \\
 & IS3 & Faulty knot technique \\
 & IS4 & Manual thread manipulation \\
 & IS5 & Result incorrect \\
\bottomrule
\end{tabular}
\caption{Common error taxonomy curated by surgical educators. Each category denotes a clinically relevant technical deviation observed during incision or suturing.}
\label{tab:incorrect_gestures}
\end{table}
\vspace{-2pt}
\subsection{Dataset Statistics}

\begin{figure*}[t]
  \centering
  \includegraphics[width=0.7\textwidth]{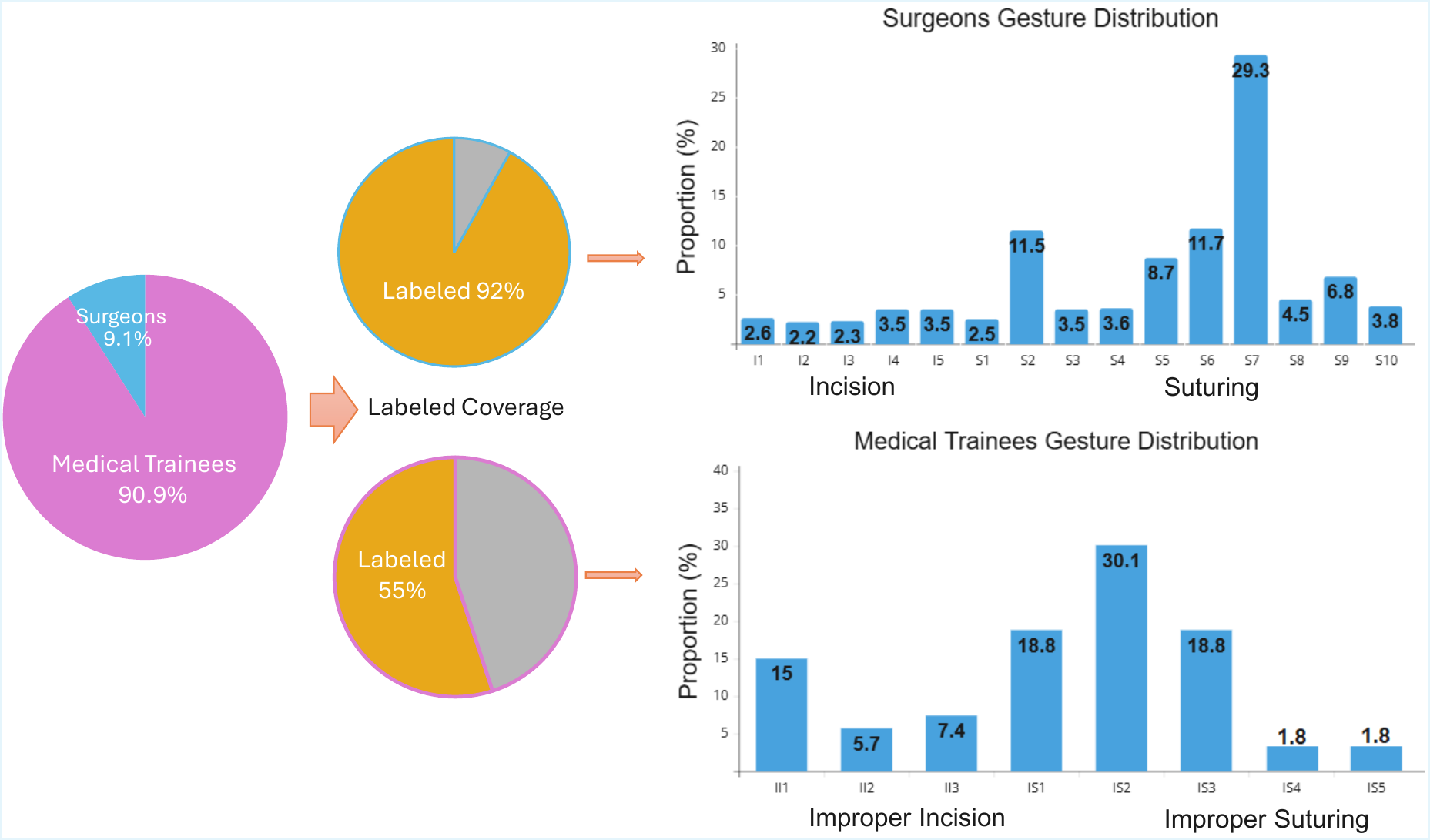}
  \caption{Overview of dataset composition and annotation coverage. The pie chart on the left illustrates the distribution of total recording time contributed by medical trainees (90.9\%) and surgeons (9.1\%). The middle plots report the proportion of annotated footage, showing 92\% labeled coverage for surgeons and 55\% for trainees. The bar plots on the right present gesture distribution for surgeons (top) and trainees (bottom), illustrating variability across incision (I1–I5) and suturing (S1–S10) categories, as well as error types (II1–II3, IS1–IS5).}
  \label{fig:data_distribution}
\end{figure*}

SHands comprises five synchronized RGB streams per recording. Trainees contributed 90.9\% of the footage, while experts provided 9.1\%. We annotated 55\% of trainee and 92\% of expert recordings to balance data diversity with high-quality benchmarking; unlabelled sequences remain available for semi-supervised learning (Figure~\ref{fig:data_distribution}). The dataset features 15 gesture classes and 8 error types. Suturing dominates the temporal distribution due to its iterative nature, with each trial requiring exactly three knots. A dedicated No Gesture/Idle class accounts for periods of inactivity or non-meaningful motion.

Temporal Dynamics: Gestures exhibit significant duration variance, reflecting authentic skill gaps. For example, S7-knot averages $40\text{s} \pm 5.8\text{s}$ (range: 30–60s), while S8-transition lasts $3\text{s} \pm 0.8\text{s}$. Total: $\sim$900K frames, 520K labeled (58\%). These temporal features are clinically valuable for skill assessment. Expert-defined frame-level boundaries ensure precision, while the error taxonomy captures any heterogeneous sub-motions that deviate from standard techniques. 

\section{Benchmark Evaluation}
We evaluate \textsc{SHands} on three core tasks that align directly with the requirements articulated by surgical-education professionals and supported by prior research highlighting the need for data-driven, objective, and scalable assessment methods for surgical skill training~\cite{lam2022machine,hla2025generative}. First, \emph{gesture recognition}, classifying video clips into one of 15 gesture primitives, measured by Top-1 accuracy and Macro-F1. Second, \emph{error detection}, a multi-label classification problem covering eight error categories, reported with Top-1 accuracy and per-class F1. Third, \emph{cross-view generalization}, assessing whether models can transfer to unseen camera viewpoints without re-training, thereby testing the view-invariance of learned representations.

\textbf{Data splits and protocols.}
We adopt a cross-subject evaluation, dividing the 52 participants between expert surgeons and medical trainees, ensuring that no subject in the test set appears during training. In single-view experiments, one camera (C1–C5) is used for both training and testing. In multi-view experiments, all five synchronized cameras are available during both phases. In cross-view generalization, models are trained on C1–C3 and evaluated on C4–C5 to test robustness to unseen viewpoints. For error detection, evaluation focuses on the trainee subset, where error frequency and clinical relevance are highest. This protocol enables a comprehensive assessment of model performance under single-view, multi-view, and cross-view conditions. Train/Val/Test = 31/11/10 participants (60/20/20), with expert/trainee breakdown: Train (12E/19T), Val (4E/7T), Test (4E/6T).

\textbf{Baseline methods.}
We benchmark a diverse set of state-of-the-art video recognition architectures covering both convolutional and transformer paradigms. For single-view gesture recognition, we include convolutional backbones R3D~\cite{tran2018closer}, SlowFast~\cite{feichtenhofer2019slowfast}, and X3D-M~\cite{feichtenhofer2020x3d}, alongside transformer-based models TimeSformer~\cite{bertasius2021timesformer}, ViViT-B~\cite{arnab2021vivit}, MViTv2-B~\cite{li2021improved}, and VideoMAE in both base and large configurations~\cite{wang2023videomaev2}. All single-view models are pretrained on Kinetics-400 or ImageNet-21K before fine-tuning on \textsc{SHands}. For multi-view learning, we evaluate architectures that explicitly capture cross-view dependencies, including MVAction~\cite{vyas2020multi} (view-specific and shared pathways), ViewCLR~\cite{das2023viewclr} (cross-view contrastive objectives), ViewCon~\cite{shah2023multi} (view-consensus fusion), and DVANet~\cite{siddiqui2024dvanet} (disentangled view-invariant representations). These multi-view methods are pretrained on NTU RGB+D and subsequently adapted for surgical gesture on \textsc{SHands}.

\vspace{-2pt}
\section{Results}
We assess \textsc{SHands} across single-view and multi-view gesture recognition, error detection, and cross-view generalization. Overall, transformer-based models outperform convolutional baselines in the single-view setting (Table~\ref{tab:sv_cls}), while multi-view methods that explicitly model cross-view dependencies deliver consistent gains (Table~\ref{tab:mv_cls}). On the trainee subset, multi-view fusion also improves error detection (Table~\ref{tab:error_det}). Cross-view transfer experiments show that DVANet achieves the best performance on unseen cameras (Table~\ref{tab:cross_view}), and a boundary-focused analysis corroborates the precision of our frame-level annotations (Figure~\ref{fig:quality}).

\subsection{Single-View Gesture Recognition}
Table~\ref{tab:sv_cls} summarizes single-view baselines. Among CNNs, SlowFast achieves 55.7\%, followed by X3D-M (53.9\%) and R3D (52.3\%). Transformers consistently outperform CNNs: TimeSformer reaches 58.4\%, ViViT-B 59.2\%, and MViTv2-B 61.8\%; masked pretraining improves performance further with VideoMAE-B (63.4\%), VideoMAE-L (65.9\%), and InternVideo2 (68.9\%). These moderate accuracies, despite strong backbones, highlight the difficulty of fine-grained tool–tissue interactions and the domain gap with generic action datasets.

\subsection{Multi-View Gesture Recognition}
Using all five synchronized views yields substantial gains (Table~\ref{tab:mv_cls}). Pretrained on NTU RGB+D 120 and fine-tuned on \textsc{SHands}, MVAction achieves Macro-F1 0.704, ViewCLR 0.717, and ViewCon 0.723. DVANet achieves 0.736 by disentangling view-specific and view-invariant factors. These results confirm that leveraging complementary viewpoints is critical for robust recognition under occlusions and viewpoint changes.

\subsection{Error Detection Performance}
To evaluate the eight error categories, we train models on \textsc{SHands}' \emph{trainee subset} using a VideoMAE-B backbone and vary the number of synchronized cameras (Table~\ref{tab:error_det}). Multi-view configurations consistently yield higher accuracy than single-view setups: the single-view VideoMAE-B reaches 60.4\%, while ViewCon improves to 66.2\% and DVANet achieves the highest accuracy of 68.5\% using all five views. Per-class results for DVANet indicate strong performance on \emph{Improper instrument grip} (77.3\%) and \emph{Incorrect result} (76.2\%), whereas more nuanced categories such as \emph{Incorrect needle handling} (60.9\%) and \emph{Excessive force application} (61.8\%) remain challenging. These findings suggest that while cross-view fusion improves robustness, recognizing subtle procedural errors still requires refined temporal and contextual modeling.

\subsection{Cross-View Generalization}
Table~\ref{tab:cross_view} analyzes the challenging cross-view transfer setting, where models trained on cameras C1–C3 are evaluated on unseen viewpoints C4–C5. On the training views, ViewCon achieves 68.4\% (Macro-F1: 0.662), and DVANet reaches 72.8\% (Macro-F1: 0.706). When tested on held-out cameras, ViewCon drops to 63.2\% (92.4\% retention), whereas DVANet remains more stable with 68.5\% (94.1\% retention). These results highlight DVANet’s superior cross-view generalization, confirming its ability to learn view-invariant representations that preserve performance across novel camera placements.

Such robustness has direct practical value: surgical training centers could deploy pre-trained models under different camera setups without requiring additional labeled data from those viewpoints, significantly reducing the cost of system adaptation. The remaining 5–6\% gap between seen and unseen views suggests residual camera-specific bias, motivating future research on stronger view-invariance constraints and meta-learning strategies for domain transfer.

\begin{table}[t]
\centering
\small
\setlength{\tabcolsep}{6pt}
\begin{tabular}{lcc}
\toprule
Method & Pretrain Dataset  & Top-1 (\%) \\
\midrule
\multicolumn{3}{l}{\textit{CNN-based architectures:}} \\
R3D \cite{tran2018closer} & K400  & $52.3 \pm 0.8$ \\
X3D-M \cite{feichtenhofer2020x3d} & K400  & $53.9 \pm 0.9$ \\
SlowFast \cite{feichtenhofer2019slowfast} & K400  & $\textbf{55.7} \pm 1.2$ \\
\midrule
\multicolumn{3}{l}{\textit{Transformer-based architectures:}} \\
TimeSformer\cite{bertasius2021timesformer} &  K400  & $58.4 \pm 1.1$ \\
ViViT-B \cite{arnab2021vivit} & K400  & $59.2 \pm 0.7$ \\
MViTv2-B\cite{li2021improved} & K400  & $61.8 \pm 0.9$ \\
VideoMAE-B\cite{wang2023videomaev2} & K400  & $63.4 \pm 0.6$ \\
VideoMAE-L\cite{wang2023videomaev2} & K400  & $65.9 \pm 0.8$ \\
InternVideo2\cite{Wang2025InternVideo2} & Im21K+K400  & $\textbf{68.9} \pm 0.8$ \\
\bottomrule
\end{tabular}
\caption{\textbf{Single-view gesture classification on SHands.} Models trained and evaluated on a single-view camera. K400 = Kinetics-400; Im21K = ImageNet-21K.}
\label{tab:sv_cls}
\end{table}

\begin{table}[t]
\centering
\small
\setlength{\tabcolsep}{5pt}
\begin{tabular}{lccc}
\toprule
Method & Pretrain Dataset & Top-1 (\%) & Macro-F1 \\
\midrule
\quad MVAction~\cite{vyas2020multi} &  NTU RGB 120& $70.4 \pm 0.9$ & 0.704 \\
\quad ViewClr~\cite{das2023viewclr}  &NTU RGB 120  & $71.1 \pm 0.7$ & 0.717 \\
\quad ViewCon~\cite{shah2023multi} &NTU RGB 120   & $72.5 \pm 0.6$ & 0.723 \\
\quad DVANet~\cite{siddiqui2024dvanet} & NTU RGB 120  & $\textbf{73.9} \pm 0.6$ & 0.736 \\
\bottomrule
\end{tabular}
\caption{\textbf{Multi-view gesture classification on SHands.} Models trained and evaluated with 5 synchronized cameras following a cross-subject protocol.}
\label{tab:mv_cls}
\end{table}

\begin{table}[t]
\centering
\small
\setlength{\tabcolsep}{6pt}
\begin{tabular}{lcc}
\toprule
Method & Camera Views & Top-1  \\
\midrule
VideoMAE-B & single-view & 60.4  \\
ViewCon\cite{shah2023multi} & All views & 66.2 \\
DVANet\cite{siddiqui2024dvanet} & All views  & \textbf{68.5} \\
\midrule
\multicolumn{3}{l}{\textit{Per-class accuracy of DAVNet (5 views):}} \\
II1 Improper instrument grip & & \textbf{77.3}  \\
II2 Incorrect angle & & 69.8  \\
II3 Repetitive cutting & & 66.2  \\
IS1 Incorrect needle handling & & 60.9  \\
IS2 Excessive force application & & 61.8  \\
IS3 Faulty knot technique & & 73.5  \\
IS4 Manual thread manipulation & & 65.1  \\
IS5 Result incorrect & & 76.2  \\
\bottomrule
\end{tabular}
\caption{\textbf{Error detection on SHands student subset.} The accuracy classification over eight error types.
}
\label{tab:error_det}
\end{table}

\begin{table}[t]
\centering
\small
\setlength{\tabcolsep}{6pt}
\begin{tabular}{lccc}
\toprule
Configuration & Method & Top-1 (\%) & Macro-F1 \\
\midrule
\multicolumn{4}{l}{\textit{Train: C1 + C2 + C3, Test: Same views}} \\
& ViewCon & 68.4 & 0.662 \\
& DVANet & \textbf{72.8} & 0.706 \\
\midrule
\multicolumn{4}{l}{\textit{Train: C1 + C2 + C3, Test: C4 + C5 (unseen)}} \\
& ViewCon & 63.2 & 0.613 \\
& DVANet & \textbf{68.5} & 0.651 \\
\midrule
\multicolumn{2}{l}{\textit{Performance retention (DVANet)}} & \textit{94.1\%} & \textit{92.2\%} \\
\bottomrule
\end{tabular}
\caption{\textbf{Cross-view generalization on SHands.} Models trained on cameras C1–C3 and evaluated on seen views (same cameras) versus unseen views (C4–C5). Retention ratio measures the preservation of performance across novel viewpoints.}
\label{tab:cross_view}
\end{table}

\begin{figure*}[t]
  \centering
  \includegraphics[width=0.66\linewidth]{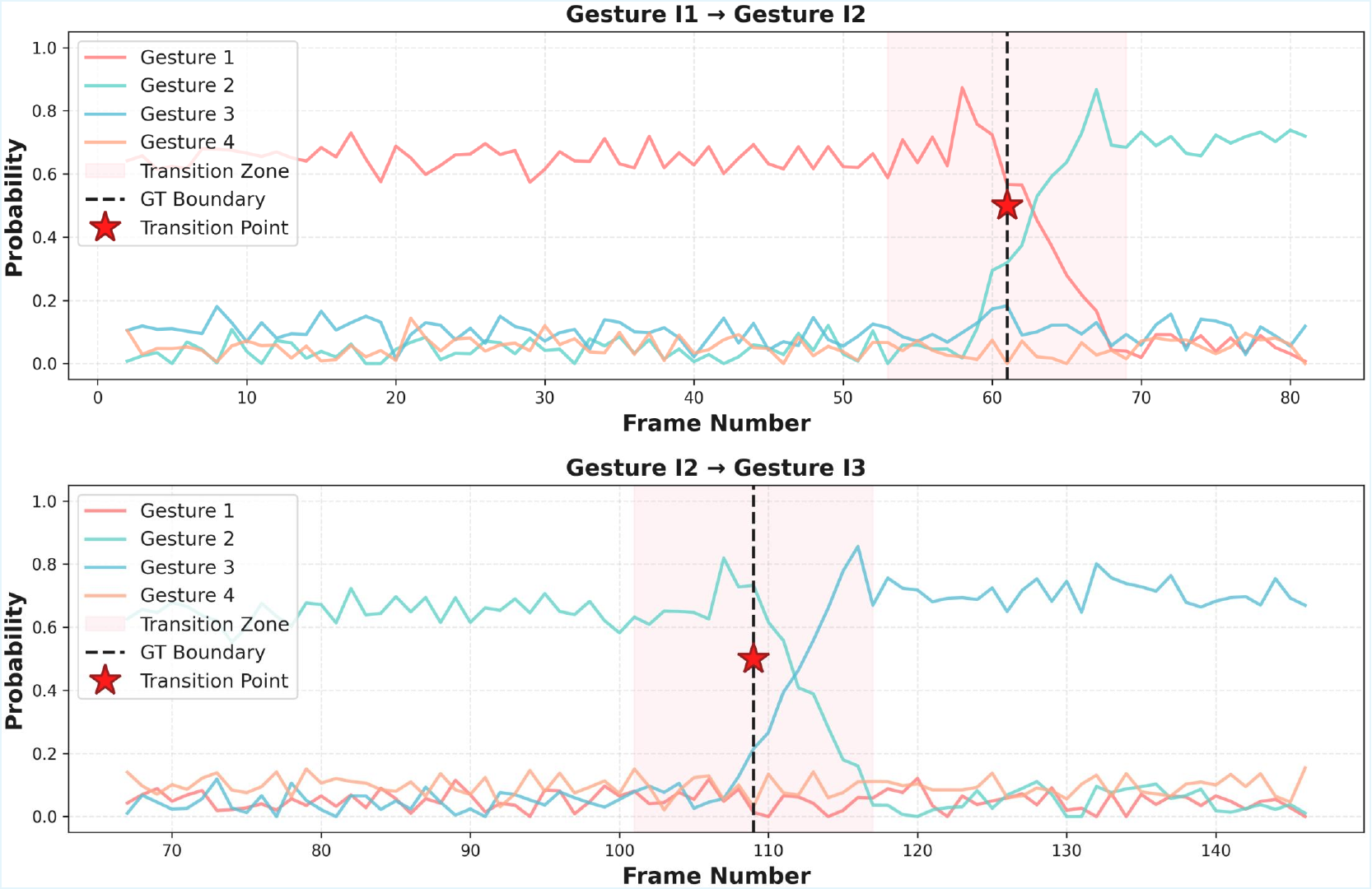}
  \caption{\textbf{Annotation quality analysis.} Gesture probability distributions predicted by DVANet around two annotated boundary frames (61 and 109). Sharp probability transitions and narrow uncertainty regions indicate high temporal precision and consistency of the manual annotations, with strong classification confidence on either side of each gesture change.}
  \label{fig:quality}
\end{figure*}

\textbf{Quality Analysis.} To assess the precision of our temporal annotations, we analyze the gesture probability distributions predicted by the multi-view DVANet model in temporal neighborhoods surrounding ground-truth boundaries. Figure~\ref{fig:quality} illustrates two representative transition points (frames 61 and 109), where sharp probability shifts between gesture classes validate the accuracy of manual annotations. The shaded regions represent uncertainty windows, while the before/after probability measures (computed over 10- and 15-frame intervals, respectively) quantify the model’s confidence on either side of each transition.
The clear separation between dominant gesture probabilities confirms that \textsc{SHands} provides temporally precise and unambiguous boundaries, enabling robust model training and reliable evaluation of segmentation accuracy.

\section{Discussion \& Future Work}
The findings above highlight the strengths and design rationale of \textsc{SHands} as a practical and pedagogically grounded benchmark for assessing surgical skill. Our decision to capture RGB video from five time-synchronized cameras was driven by educational constraints and by the need for realistic deployment. In both formal training facilities and personal study environments, standard webcams and smartphones provide an affordable and easy-to-deploy setup for recording training sessions. 

By focusing on RGB only, we emphasize an \emph{anytime, anywhere} training paradigm in which models can be deployed with a single camera while still benefiting from knowledge distilled from multi-view ``teacher'' systems. We note that although depth or force sensors can provide additional geometric/interaction cues, adopting these modalities requires significantly more complex hardware integration and precise multi-sensor calibration. Depth systems in particular are sensitive to occlusions and reflective surgical instruments, making them unreliable in realistic training environments. These limitations render such sensor-dependent setups impractical for routine use in clinical training for medical trainees.

Moreover, our focus on linear incision and suturing reflects their foundational role in surgical curricula and their ubiquity in skills-lab training. The resulting taxonomy of 15 gesture primitives captures the key hand–tool configurations identified by the project’s medical educators. At the same time, the eight error categories were selected for their direct observability in RGB and instructional actionability, capturing information about grip stability, approach angle, tissue control, force application, and knot security.

Looking ahead, we will extend our approach toward holistic skill assessment. This involves aggregating temporal gesture predictions into interpretable global ratings (e.g., excellent, good, average, poor) with calibrated confidence and pairwise ranking between expert and trainee sessions. Such modeling opens the path toward a continuous feedback loop in which students practice in physical or VR simulators, a low-latency recognizer detects gestures and errors in real time, and a language model translates these detections into actionable, context-aware guidance. The proposed teacher–student distillation paradigm (multi-view teacher $\rightarrow$ single-view student) aligns naturally with single-camera or head-mounted configurations, enabling scalable, low-infrastructure deployment across training institutions.

Finally, \textsc{SHands} highlights the intrinsic complexity of real surgical behavior. Gesture imbalance and ambiguous temporal boundaries arise not from annotation errors but from the intrinsic variability of human motion. Rather than simplifying this variability, our dataset preserves it to reflect realistic conditions. Future work can address these challenges through temporal uncertainty modeling, curriculum-aware sampling strategies, and active annotation protocols that refine label consistency without compromising realism.

\section{Conclusion}
This paper presented \textsc{Open-Surgery Hands} (\textsc{SHands}), a synchronized multi-view dataset and benchmark for open-surgical gesture and error recognition. The dataset encompasses two fundamental tasks, incision and suturing, decomposed into 15 gesture primitives and eight clinically validated error categories. Experimental analyses demonstrate that multi-view supervision significantly enhances robustness and boundary precision, while RGB-only sensing enables realistic and privacy-preserving deployment for surgical training. \textsc{SHands} establishes a foundation for data-driven, interpretable, and scalable assessment of technical surgical skills, bridging the gap between computer vision research and medical education.

\section{Ethics Statement}
All data collection procedures were approved by the institutional ethics committee, and written informed consent was obtained from the participants. Recordings captured only hands and forearms, with no identifiable personal information. Participants were informed of the research objectives and the intended use of the data for training AI models. The dataset will be released under a Creative Commons Attribution 4.0 International License, permitting academic and commercial use while prohibiting any application involving surveillance, biometric identification, or non-consensual monitoring.

\section*{Acknowledgments}
Supported by IDS (100.133 IP-ICT) and INDUX-R (GA No. 101135556; DOI: 10.3030/101135556). Funded by the European Union and the Swiss State Secretariat for Education, Research and Innovation (SERI).
Disclaimer: Opinions expressed are the authors' alone and do not necessarily represent the EU or CINEA. Neither the EU nor the granting authority is responsible for the content.

\bibliographystyle{IEEEtran}
\bibliography{egbib} 

@String(CVPR= {IEEE Conf. Comput. Vis. Pattern Recog.})

@String(ICCV= {Int. Conf. Comput. Vis.})

@String(ECCV= {Eur. Conf. Comput. Vis.})

@String(BMVC= {Brit. Mach. Vis. Conf.})

@String(AAAI = {AAAI})

@String(CVPR  = {CVPR})

@String(ICCV  = {ICCV})

@String(ECCV  = {ECCV})

@String(BMVC  =	{BMVC})

@inproceedings{ma2024transsg,
  title={TransSG: a spatial-temporal transformer for surgical gesture recognition},
  author={Ma, Le and Kang, Hangyeol and Magnenat-Thalmann, Nadia and Wac, Katarzyna},
  booktitle={Computer Graphics International Conference},
  pages={151--165},
  year={2024},
  organization={Springer}
}

@inproceedings{farha2019ms,
  title={Ms-tcn: Multi-stage temporal convolutional network for action segmentation},
  author={Farha, Yazan Abu and Gall, Jurgen},
  booktitle={Proceedings of the IEEE Conference on Computer Vision and Pattern Recognition},
  pages={3575--3584},
  year={2019}
}

@inproceedings{gao2014jhu,
  author    = {Yixin Gao and S. Swaroop Vedula and Carol E. Reiley and Narges Ahmidi and Balakrishnan Varadarajan and Henry C. Lin and Lingling Tao and Luca Zappella and Benjam{\'\i}n B{\'e}jar and David D. Yuh and Chi Chiung Grace Chen and Ren{\'e} Vidal and Sanjeev Khudanpur and Gregory D. Hager},
  title     = {JHU-ISI Gesture and Skill Assessment Working Set (JIGSAWS): A Surgical Activity Dataset for Human Motion Modeling},
  booktitle = {Modeling and Monitoring of Computer Assisted Interventions (M2CAI) -- MICCAI Workshop},
  volume    = {3},
  pages     = {3},
  year      = {2014}
}

@article{twinanda2017endonet,
  author  = {Andru P. Twinanda and Sherif Shehata and Didier Mutter and Jacques Marescaux and Michel de Mathelin and Nicolas Padoy},
  title   = {EndoNet: A Deep Architecture for Recognition Tasks on Laparoscopic Videos},
  journal = {IEEE Transactions on Medical Imaging},
  volume  = {36},
  number  = {1},
  pages   = {86--97},
  year    = {2017},
  doi     = {10.1109/TMI.2016.2593957}
}

@article{jin2018sv,
  author  = {Yueming Jin and Qi Dou and Hao Chen and Lequan Yu and Jing Qin and Chi-Wing Fu and Pheng-Ann Heng},
  title   = {SV-RCNet: Workflow Recognition from Surgical Videos Using Recurrent Convolutional Network},
  journal = {IEEE Transactions on Medical Imaging},
  volume  = {37},
  number  = {5},
  pages   = {1114--1126},
  year    = {2018}
}

@inproceedings{shahroudy2016ntu,
  author    = {Amir Shahroudy and Jun Liu and Tian-Tsong Ng and Gang Wang},
  title     = {{NTU RGB+D}: A Large Scale Dataset for {3D} Human Activity Analysis},
  booktitle = {Proceedings of the IEEE Conference on Computer Vision and Pattern Recognition (CVPR)},
  pages     = {1010--1019},
  year      = {2016}
}

@inproceedings{sharghi2020operatingroom,
  author    = {Aria P. Sharghi and Alex Haugerud and Dixon Oh and Nima Mohareri},
  title     = {Automatic Operating Room Surgical Activity Recognition for Robot-Assisted Surgery},
  booktitle = {International Conference on Medical Image Computing and Computer-Assisted Intervention (MICCAI)},
  series    = {Lecture Notes in Computer Science},
  volume    = {12263},
  pages     = {385--395},
  publisher = {Springer},
  year      = {2020}
}

@article{alhajj2019cataracts,
  author  = {Hassan Al Hajj and Mathieu Lamard and Pierre{-}Henri Conze and B{\'e}atrice Cochener and G{\'e}rard Quellec},
  title   = {{CATARACTS}: Challenge on Automatic Tool Annotation for Cataract Surgery},
  journal = {Medical Image Analysis},
  volume  = {52},
  pages   = {24--41},
  year    = {2019},
  doi     = {10.1016/j.media.2018.11.008}
}

@article{malpani2016systemevents,
  author  = {Anand Malpani and Colin Lea and Chi Chiung Grace Chen and Gregory D. Hager},
  title   = {System events: readily accessible features for surgical phase detection},
  journal = {International Journal of Computer Assisted Radiology and Surgery},
  volume  = {11},
  number  = {6},
  pages   = {1201--1209},
  year    = {2016}
}

@article{bai2023surgicalvqla,
  author  = {Long Bai and Mobarakol Islam and Lalithkumar Seenivasan and Hongliang Ren},
  title   = {Surgical-VQLA: Transformer with Gated Vision-Language Embedding for Visual Question Localized-Answering in Robotic Surgery},
  journal = {arXiv preprint arXiv:2305.11692},
  year    = {2023}
}

@article{liu2017pku,
  author  = {Chunhui Liu and Yueyu Hu and Yanghao Li and Sijie Song and Jiaying Liu},
  title   = {PKU-MMD: A Large Scale Benchmark for Continuous Multi-Modal Human Action Understanding},
  journal = {arXiv preprint arXiv:1703.07475},
  year    = {2017}
}

@inproceedings{kwon2021h2o,
  author    = {Taein Kwon and Bugra Tekin and Jan St{\"u}hmer and Federica Bogo and Marc Pollefeys},
  title     = {H2O: Two Hands Manipulating Objects for First Person Interaction Recognition},
  booktitle = {Proceedings of the IEEE/CVF International Conference on Computer Vision (ICCV)},
  pages     = {10138--10148},
  year      = {2021}
}

@inproceedings{sener2022assembly101,
  author    = {Fadime Sener and Dibyadip Chatterjee and Daniel Shelepov and Kun He and Dipika Singhania and Robert Wang and Angela Yao},
  title     = {Assembly101: A Large-Scale Multi-View Video Dataset for Understanding Procedural Activities},
  booktitle = {Proceedings of the IEEE/CVF Conference on Computer Vision and Pattern Recognition (CVPR)},
  pages     = {21066--21076},
  year      = {2022}
}

@inproceedings{yi2021asformer,
  author    = {Fangqiu Yi and Hongyu Wen and Tingting Jiang},
  title     = {ASFormer: Transformer for Action Segmentation},
  booktitle = {British Machine Vision Conference (BMVC)},
  year      = {2021}
}

@article{li2020mstcnpp,
  author  = {Shi{-}Jie Li and Yazan AbuFarha and Yun Liu and Ming{-}Ming Cheng and J{\"u}rgen Gall},
  title   = {{MS\text{-}TCN++}: Multi-Stage Temporal Convolutional Network for Action Segmentation},
  journal = {IEEE Transactions on Pattern Analysis and Machine Intelligence},
  year    = {2020},
  doi     = {10.1109/TPAMI.2020.3021756}
}

@inproceedings{tran2018closer,
  author    = {Du Tran and Heng Wang and Lorenzo Torresani and Jamie Ray and Yann LeCun and Manohar Paluri},
  title     = {A Closer Look at Spatiotemporal Convolutions for Action Recognition},
  booktitle = {Proceedings of the IEEE/CVF Conference on Computer Vision and Pattern Recognition (CVPR)},
  pages     = {6450--6459},
  year      = {2018}
}

@inproceedings{tong2022videomae,
  author    = {Zhan Tong and Yibing Song and Jue Wang and Limin Wang},
  title     = {VideoMAE: Masked Autoencoders are Data-Efficient Learners for Self-Supervised Video Pre-Training},
  booktitle = {Advances in Neural Information Processing Systems (NeurIPS)},
  year      = {2022}
}

@inproceedings{wang2023videomaev2,
  author    = {Limin Wang and Bingkun Huang and Zhiyu Zhao and Zhan Tong and Yinan He and Yi Wang and Yali Wang and Yu Qiao},
  title     = {VideoMAE V2: Scaling Video Masked Autoencoders with Dual Masking},
  booktitle = {Proceedings of the IEEE/CVF Conference on Computer Vision and Pattern Recognition (CVPR)},
  year      = {2023}
}

@inproceedings{grauman2024egoexo4d,
  title={Ego-exo4d: Understanding skilled human activity from first-and third-person perspectives},
  author={Grauman, Kristen and Westbury, Andrew and Torresani, Lorenzo and Kitani, Kris and Malik, Jitendra and Afouras, Triantafyllos and Ashutosh, Kumar and Baiyya, Vijay and Bansal, Siddhant and Boote, Bikram and others},
  booktitle={Proceedings of the IEEE/CVF Conference on Computer Vision and Pattern Recognition},
  pages={19383--19400},
  year={2024}
}

@inproceedings{huang2024egoexolearn,
  author = {Huang, Yifei and Chen, Guo and Xu, Jilan and Zhang, Mingfang and Yang, Lijin and Pei, Baoqi and Zhang, Hongjie and Dong, Lu and Wang, Yali and Wang, Limin and Qiao, Yu},
  title = {EgoExoLearn: A Dataset for Bridging Asynchronous Ego- and Exo-centric View of Procedural Activities in Real World},
  booktitle = {CVPR},
  year = {2024}
}

@inproceedings{ghoddoosian2023ata,
  author = {Ghoddoosian, Reza and Dwivedi, Isha and Agarwal, Naman and Dariush, Bo},
  title = {Weakly-Supervised Action Segmentation and Unseen Error Detection in Anomalous Instructional Videos},
  booktitle = {ICCV},
  year = {2023},
  note = {Introduces the ATA dataset}
}

@inproceedings{wang2023holoassist,
  title={Holoassist: an egocentric human interaction dataset for interactive ai assistants in the real world},
  author={Wang, Xin and Kwon, Taein and Rad, Mahdi and Pan, Bowen and Chakraborty, Ishani and Andrist, Sean and Bohus, Dan and Feniello, Ashley and Tekin, Bugra and Frujeri, Felipe Vieira and others},
  booktitle={Proceedings of the IEEE/CVF International Conference on Computer Vision},
  pages={20270--20281},
  year={2023}
}

@inproceedings{zhang2023logo,
  author = {Zhang, Shiyi and Dai, Wenxun and Wang, Sujia and Shen, Xiangwei and Lu, Jiwen and Zhou, Jie and Tang, Yansong},
  title = {LOGO: A Long-Form Video Dataset for Group Action Quality Assessment},
  booktitle = {CVPR},
  year = {2023}
}

@article{liu2025lovit,
  title={Lovit: Long video transformer for surgical phase recognition},
  author={Liu, Yang and Boels, Maxence and Garcia-Peraza-Herrera, Luis C and Vercauteren, Tom and Dasgupta, Prokar and Granados, Alejandro and Ourselin, Sebastien},
  journal={Medical Image Analysis},
  volume={99},
  pages={103366},
  year={2025},
  publisher={Elsevier}
}

@inproceedings{fujii2024egosurgery,
  title={Egosurgery-phase: a dataset of surgical phase recognition from egocentric open surgery videos},
  author={Fujii, Ryo and Hatano, Masashi and Saito, Hideo and Kajita, Hiroki},
  booktitle={International Conference on Medical Image Computing and Computer-Assisted Intervention},
  pages={187--196},
  year={2024},
  organization={Springer}
}

@inproceedings{weerasinghe2024multimodal,
  title={Multimodal transformers for real-time surgical activity prediction},
  author={Weerasinghe, Keshara and Roodabeh, Seyed Hamid Reza and Hutchinson, Kay and Alemzadeh, Homa},
  booktitle={2024 IEEE International Conference on Robotics and Automation (ICRA)},
  pages={13323--13330},
  year={2024},
  organization={IEEE}
}

@inproceedings{wang2024cpr,
  title={CPR-Coach: Recognizing composite error actions based on single-class training},
  author={Wang, Shunli and Wang, Shuaibing and Yang, Dingkang and Li, Mingcheng and Kuang, Haopeng and Zhao, Xiao and Su, Liuzhen and Zhai, Peng and Zhang, Lihua},
  booktitle={Proceedings of the IEEE/CVF Conference on Computer Vision and Pattern Recognition},
  pages={18782--18792},
  year={2024}
}

@inproceedings{vyas2020multi,
  title={Multi-view action recognition using cross-view video prediction},
  author={Vyas, Shruti and Rawat, Yogesh S and Shah, Mubarak},
  booktitle={European Conference on Computer Vision},
  pages={427--444},
  year={2020},
  organization={Springer}
}

@inproceedings{liu2023skit,
  title={Skit: a fast key information video transformer for online surgical phase recognition},
  author={Liu, Yang and Huo, Jiayu and Peng, Jingjing and Sparks, Rachel and Dasgupta, Prokar and Granados, Alejandro and Ourselin, Sebastien},
  booktitle={Proceedings of the IEEE/CVF international conference on computer vision},
  pages={21074--21084},
  year={2023}
}

@inproceedings{yang2023vidchapters7m,
  title={Vidchapters-7m: Video chapters at scale},
  author={Yang, Antoine and Nagrani, Arsha and Laptev, Ivan and Sivic, Josef and Schmid, Cordelia},
  booktitle={Advances in Neural Information Processing Systems},
  volume={36},
  pages={49428--49444},
  year={2023}
}

@inproceedings{tang2023m3net,
  title={M3net: multi-view encoding, matching, and fusion for few-shot fine-grained action recognition},
  author={Tang, Hao and Liu, Jun and Yan, Shuanglin and Yan, Rui and Li, Zechao and Tang, Jinhui},
  booktitle={Proceedings of the 31st ACM international conference on multimedia},
  pages={1719--1728},
  year={2023}
}

@article{jamal2023surgmae,
  title={SurgMAE: Masked autoencoders for long surgical video analysis},
  author={Jamal, Muhammad Abdullah and Mohareri, Omid},
  journal={arXiv preprint arXiv:2305.11451},
  year={2023}
}

@inproceedings{wang2024mv2mae,
  title={Mv2mae: Multi-view video masked autoencoders},
  author={Shah, Ketul and Crandall, Robert and Xu, Jie and Zhou, Peng and George, Marian and Bansal, Mayank and Chellappa, Rama},
  journal={arXiv preprint arXiv:2401.15900},
  year={2024}
}

@inproceedings{li2021improved,
  title={MViTv2: Improved multiscale vision transformers for classification and detection},
  author={Li, Yanghao and Wu, Chao-Yuan and Fan, Haoqi and Mangalam, Karttikeya and Xiong, Bo and Malik, Jitendra and Feichtenhofer, Christoph},
  booktitle={CVPR},
  year={2022}
}

@inproceedings{das2023viewclr,
  title={Viewclr: Learning self-supervised video representation for unseen viewpoints},
  author={Das, Srijan and Ryoo, Michael S},
  booktitle={Proceedings of the IEEE/CVF Winter Conference on Applications of Computer Vision},
  pages={5573--5583},
  year={2023}
}

@inproceedings{feichtenhofer2020x3d,
  title={X3d: Expanding architectures for efficient video recognition},
  author={Feichtenhofer, Christoph},
  booktitle={Proceedings of the IEEE/CVF conference on computer vision and pattern recognition},
  pages={203--213},
  year={2020}
}

@inproceedings{arnab2021vivit,
  title={Vivit: A video vision transformer},
  author={Arnab, Anurag and Dehghani, Mostafa and Heigold, Georg and Sun, Chen and Lu{\v{c}}i{\'c}, Mario and Schmid, Cordelia},
  booktitle={Proceedings of the IEEE/CVF international conference on computer vision},
  pages={6836--6846},
  year={2021}
}

@inproceedings{shah2023multi,
  title={Multi-view action recognition using contrastive learning},
  author={Shah, Ketul and Shah, Anshul and Lau, Chun Pong and de Melo, Celso M and Chellappa, Rama},
  booktitle={Proceedings of the IEEE/CVF Winter Conference on Applications of Computer Vision},
  pages={3381--3391},
  year={2023}
}

@article{hu2024multi,
  title={Multi-view masked contrastive representation learning for endoscopic video analysis},
  author={Hu, Kai and Xiao, Ye and Zhang, Yuan and Gao, Xieping},
  journal={Advances in Neural Information Processing Systems},
  volume={37},
  pages={47987--48014},
  year={2024}
}

@inproceedings{bertasius2021timesformer,
  author    = {Bertasius, Gedas and Wang, Heng and Torresani, Lorenzo},
  title     = {Is Space-Time Attention All You Need for Video Understanding?},
  booktitle = {Proceedings of the 38th International Conference on Machine Learning (ICML)},
  year      = {2021},
  series    = {PMLR},
  volume    = {139}
}

@inproceedings{feichtenhofer2019slowfast,
  author    = {Feichtenhofer, Christoph and Fan, Haoqi and Malik, Jitendra and He, Kaiming},
  title     = {SlowFast Networks for Video Recognition},
  booktitle = {Proceedings of the IEEE/CVF International Conference on Computer Vision (ICCV)},
  year      = {2019},
  pages     = {6202--6211}
}

@inproceedings{fan2021mvit,
  author    = {Fan, Haoqi and Xiong, Bo and Mangalam, Karttikeya and Li, Yanghao and Yan, Zhicheng and Malik, Jitendra and Feichtenhofer, Christoph},
  title     = {Multiscale Vision Transformers},
  booktitle = {Proceedings of the IEEE/CVF International Conference on Computer Vision (ICCV)},
  year      = {2021}
}

@inproceedings{wang2023endofm,
  author    = {Wang, Zhao and Liu, Chang and Zhang, Shaoting and Dou, Qi},
  title     = {Foundation Model for Endoscopy Video Analysis via Large-scale Self-supervised Pre-train},
  booktitle = {Medical Image Computing and Computer Assisted Intervention – MICCAI 2023},
  series    = {Lecture Notes in Computer Science},
  volume    = {14223},
  pages     = {101--111},
  publisher = {Springer},
  year      = {2023},
  doi       = {10.1007/978-3-031-43996-4_10}
}

@article{batic2024endovit,
  title={Endovit: pretraining vision transformers on a large collection of endoscopic images},
  author={Bati{\'c}, Dominik and Holm, Felix and {\"O}zsoy, Ege and Czempiel, Tobias and Navab, Nassir},
  journal={International Journal of Computer Assisted Radiology and Surgery},
  volume={19},
  number={6},
  pages={1085--1091},
  year={2024},
  publisher={Springer}
}

@inproceedings{siddiqui2024dvanet,
  author    = {Siddiqui, Yawar and Shaik, Mohammed Azharuddin and             Song, Limin and Nie{\ss}ner, Matthias and Dai, Angela},
  title     = {{DVANet}: Disentangling View and Action Features for Multiview Action Recognition},
  booktitle = {Proceedings of the AAAI Conference on Artificial Intelligence (AAAI)},
  year      = {2024},
  volume    = {38},
  number    = {5},
  doi       = {10.1609/aaai.v38i5.28290}
}

@inproceedings{qian2021cvrl,
  author    = {Qian, Rui and Targarona, Jo{\'a}n and {\"O}zdenizci, Ozan and Echevarria, Juan and Shen, Yandong and Liu, Chang and Sclaroff, Stan and Saenko, Kate and Wu, Bryan and Feris, Rogerio},
  title     = {Spatiotemporal Contrastive Video Representation Learning},
  booktitle = {Proceedings of the IEEE/CVF Conference on Computer Vision and Pattern Recognition (CVPR)},
  year      = {2021}
}

@inproceedings{madapana2019desk,
  title={Desk: A robotic activity dataset for dexterous surgical skills transfer to medical robots},
  author={Madapana, Naveen and Rahman, Md Masudur and Sanchez-Tamayo, Natalia and Balakuntala, Mythra V and Gonzalez, Glebys and Bindu, Jyothsna Padmakumar and Venkatesh, LN Vishnunandan and Zhang, Xingguang and Noguera, Juan Barragan and Low, Thomas and others},
  booktitle={2019 IEEE/RSJ International Conference on Intelligent Robots and Systems (IROS)},
  pages={6928--6934},
  year={2019},
  organization={IEEE}
}

@inproceedings{wang2022autolaparo,
  title     = {AutoLaparo: A New Dataset of Integrated Multi-tasks for Image-guided Surgical Automation in Laparoscopic Hysterectomy},
  author    = {Wang, Ziyi and Lu, Bo and Long, Yonghao and Zhong, Fangxun and Cheung, Tak-Hong and Dou, Qi and Liu, Yunhui},
  booktitle = {Medical Image Computing and Computer-Assisted Intervention -- MICCAI 2022},
  series    = {Lecture Notes in Computer Science},
  pages     = {486--496},
  year      = {2022},
  publisher = {Springer},
  doi       = {10.1007/978-3-031-16449-1_46}
}

@inproceedings{ozsoy2022_4dor,
  title     = {4D-OR: Semantic Scene Graphs for OR Domain Modeling},
  author    = {{\"O}zsoy, Ege and {\"O}rnek, Evin P{\i}nar and Eck, Ulrich and Czempiel, Tobias and Tombari, Federico and Navab, Nassir},
  booktitle = {Medical Image Computing and Computer-Assisted Intervention -- MICCAI 2022},
  series    = {Lecture Notes in Computer Science},
  volume    = {13437},
  pages     = {475--485},
  year      = {2022},
  publisher = {Springer},
  doi       = {10.1007/978-3-031-16449-1_45}
}

@inproceedings{rivas2023surgical,
  title={A surgical dataset from the da Vinci Research Kit for task automation and recognition},
  author={Rivas-Blanco, Irene and Del-Pulgar, Carlos J P{\'e}rez and Mariani, Andrea and Tortora, Giuseppe and Reina, Antonio J},
  booktitle={2023 3rd International Conference on Electrical, Computer, Communications and Mechatronics Engineering (ICECCME)},
  pages={1--6},
  year={2023},
  organization={IEEE}
}

@article{liu2019ntu,
  title={Ntu rgb+ d 120: A large-scale benchmark for 3d human activity understanding},
  author={Liu, Jun and Shahroudy, Amir and Perez, Mauricio and Wang, Gang and Duan, Ling-Yu and Kot, Alex C},
  journal={IEEE transactions on pattern analysis and machine intelligence},
  volume={42},
  number={10},
  pages={2684--2701},
  year={2019},
  publisher={IEEE}
}

@inproceedings{Parmar2019MTLAQA,
  title     = {What and How Well You Performed? A Multitask Learning Approach to Action Quality Assessment},
  author    = {Parmar, Paritosh and Morris, Brendan Tran},
  booktitle = {Proceedings of the IEEE/CVF Conference on Computer Vision and Pattern Recognition (CVPR)},
  pages     = {304--313},
  year      = {2019}
}

@inproceedings{Xu2022FineDiving,
  title     = {FineDiving: A Fine-grained Dataset for Procedure-aware Action Quality Assessment},
  author    = {Xu, Jinglin and Rao, Yongming and Yu, Xumin and Chen, Guangyi and Zhou, Jie and Lu, Jiwen},
  booktitle = {Proceedings of the IEEE/CVF Conference on Computer Vision and Pattern Recognition (CVPR)},
  pages     = {2949--2958},
  year      = {2022}
}

@inproceedings{srivastav2018mvor,
  title={MVOR: A Multi-view RGB-D Operating Room Dataset for 2D and 3D Human Pose Estimation},
  author={Srivastav, Vinkle Kumar and Issenhuth, Thibaut and Kadkhodamohammadi, Abdolrahim and de Mathelin, Michel and Gangi, Afshin and Padoy, Nicolas},
  booktitle={MICCAI 2018 Satellite Workshop, Granada, Spain, september 16-20 2018},
  year={2018},
  organization={Springer}
}

@inproceedings{ozsoy2025_mmor,
  title     = {MM-OR: A Large Multimodal Operating Room Dataset for Semantic Understanding of High-Intensity Surgical Environments},
  author    = {{\"O}zsoy, Ege and Pellegrini, Chantal and Czempiel, Tobias and Tristram, Felix and Yuan, Kun and Bani-Harouni, David and Eck, Ulrich and Busam, Benjamin and Keicher, Matthias and Navab, Nassir},
  booktitle = {Proceedings of the IEEE/CVF Conference on Computer Vision and Pattern Recognition (CVPR)},
  year      = {2025},
  note      = {arXiv:2503.02579}
}

@inproceedings{lin2024human,
  title={Human-Centric Transformer for Domain Adaptive Action Recognition},
  author={Lin, Kun-Yu and Zhou, Jiaming and Zheng, Wei-Shi},
  booktitle={IEEE Transactions on Pattern Analysis and Machine Intelligence},
  year={2024}
}

@inproceedings{ho2025dejavid,
  title={DejaVid: Encoder-Agnostic Learned Temporal Matching for Video Classification},
  author={Ho, Darryl and Madden, Samuel},
  booktitle={Proceedings of the Computer Vision and Pattern Recognition Conference},
  pages={24023--24032},
  year={2025}
}

@article{lin2025mv,
  title={Mv-gmn: State space model for multi-view action recognition},
  author={Lin, Yuhui and Lu, Jiaxuan and Yong, Yue and Zhang, Jiahao},
  journal={arXiv preprint arXiv:2501.13829},
  year={2025}
}

@inproceedings{Pan2025BASKET,
  title     = {BASKET: A Large-Scale Video Dataset for Fine-Grained Skill Estimation},
  author    = {Pan, Yulu and Zhang, Ce and Bertasius, Gedas},
  booktitle = {Proceedings of the IEEE/CVF Conference on Computer Vision and Pattern Recognition (CVPR)},
  year      = {2025}
}

@incollection{Wang2025InternVideo2,
  title={InternVideo2: Scaling Foundation Models for Multimodal Video Understanding},
  author={Wang, Y. and Chen, J. and Li, K. and Pan, Z. and Li, J. and Wen, Z. and Li, S. and Wang, H. and Lu, H. and Qiao, Y.},
  editor={Leonardis, A. and Ricci, E. and Roth, S. and Russakovsky, O. and Sattler, T. and Varol, G.},
  booktitle={Computer Vision -- ECCV 2024},
  series={Lecture Notes in Computer Science},
  volume={15143},
  pages={370--389},
  year={2025},
  publisher={Springer, Cham},
  doi={10.1007/978-3-031-73013-9\_23}
}

@article{martin1997objective,
  title={Objective structured assessment of technical skill (OSATS) for surgical residents},
  author={Martin, JA and Regehr, Glenn and Reznick, Richard and Macrae, Helen and Murnaghan, John and Hutchison, Carol and Brown, M},
  journal={British journal of surgery},
  volume={84},
  number={2},
  pages={273--278},
  year={1997},
  publisher={Wiley Online Library}
}

@article{shayan2023measuring,
  title={Measuring hand movement for suturing skill assessment: A simulation-based study},
  author={Shayan, Amir Mehdi and Singh, Simar and Gao, Jianxin and Groff, Richard E and Bible, Joe and Eidt, John F and Sheahan, Malachi and Gandhi, Sagar S and Blas, Joseph V and Singapogu, Ravikiran},
  journal={Surgery},
  volume={174},
  number={5},
  pages={1184--1192},
  year={2023},
  publisher={Elsevier}
}

@article{olsen2022crowdsourced,
  title={Crowdsourced assessment of surgical skills: a systematic review},
  author={Olsen, Rikke G and Gen{\'e}t, Malthe F and Konge, Lars and Bjerrum, Flemming},
  journal={The American Journal of Surgery},
  volume={224},
  number={5},
  pages={1229--1237},
  year={2022},
  publisher={Elsevier}
}

@article{olsen2025untangling,
  title={Untangling surgical gesture analysis—are we even speaking the same language? a systematic review},
  author={Olsen, Rikke Groth and Andersen, Annarita Ghosh and Hung, Andrew J and Svendsen, Morten Bo S{\o}ndergaard and Dagn{\ae}s-Hansen, Julia Abildgaard and Konge, Lars and R{\o}der, Andreas and Bjerrum, Flemming},
  journal={Surgical Endoscopy},
  pages={1--20},
  year={2025},
  publisher={Springer}
}

@article{pedrett2023technical,
  title={Technical skill assessment in minimally invasive surgery using artificial intelligence: a systematic review},
  author={Pedrett, Romina and Mascagni, Pietro and Beldi, Guido and Padoy, Nicolas and Lavanchy, Joel L},
  journal={Surgical endoscopy},
  volume={37},
  number={10},
  pages={7412--7424},
  year={2023},
  publisher={Springer}
}

@article{power2025automated,
  title={Automated assessment of simulated laparoscopic surgical skill performance using deep learning},
  author={Power, David and Burke, Cathy and Madden, Michael G and Ullah, Ihsan},
  journal={Scientific Reports},
  volume={15},
  number={1},
  pages={13591},
  year={2025},
  publisher={Nature Publishing Group UK London}
}

@article{hla2025generative,
  title={Generative AI \& machine learning in surgical education},
  author={Hla, Diana A and Hindin, David I},
  journal={Current problems in surgery},
  volume={63},
  pages={101701},
  year={2025},
  publisher={Elsevier}
}

@article{lam2022machine,
  title={Machine learning for technical skill assessment in surgery: a systematic review},
  author={Lam, Kyle and Chen, Junhong and Wang, Zeyu and Iqbal, Fahad M and Darzi, Ara and Lo, Benny and Purkayastha, Sanjay and Kinross, James M},
  journal={NPJ digital medicine},
  volume={5},
  number={1},
  pages={24},
  year={2022},
  publisher={Nature Publishing Group UK London}
}

\end{document}